# *Konooz*: Multi-domain Multi-dialect Corpus for Named Entity Recognition

**Nagham Hamad** [λ,σ]    **Mohammed Khalilia**[λ]    **Mustafa Jarrar**[λ,ξ]

[λ] Birzeit University, Palestine
[ξ] Hamad Bin Khalifa University, Qatar
[σ] Palestine Technical University-Kadoorie, Palestine

{nhamad,mkhalilia,mjarrar}@birzeit.edu

## Abstract

We introduce *Konooz*, a novel multi-dimensional corpus covering 16 Arabic dialects across 10 domains, resulting in 160 distinct corpora. The corpus comprises about 777k tokens, carefully collected and manually annotated with 21 entity types using both nested and flat annotation schemes - using the Wojood guidelines. While *Konooz* is useful for various NLP tasks like domain adaptation and transfer learning, this paper primarily focuses on benchmarking existing Arabic Named Entity Recognition (NER) models, especially cross-domain and cross-dialect model performance. Our benchmarking of four Arabic NER models using *Konooz* reveals a significant drop in performance of up to 38% when compared to the in-distribution data. Furthermore, we present an in-depth analysis of domain and dialect divergence and the impact of resource scarcity. We also measured the overlap between domains and dialects using the Maximum Mean Discrepancy (MMD) metric, and illustrated why certain NER models perform better on specific dialects and domains. *Konooz* is open-source and publicly available at https://sina.birzeit.edu/wojood/#download

## 1 Introduction

NER is crucial in various NLP tasks, including machine translation (Hassan and Sorensen, 2005; Darwish et al., 2021), word sense disambiguation (Jarrar et al., 2023b; Al-Hajj and Jarrar, 2021), data extraction (Barbon Junior et al., 2024), language understanding (Khalilia et al., 2024), question answering (Badawy et al., 2011), and interoperability (Jarrar et al., 2011). State-of-the-art Arabic NER models demonstrate impressive performance, achieving $F_1$-scores above 90% (Jarrar et al., 2024). However, these models continue to face challenges across various domains and dialects (Singhal et al., 2023). Arabic dialects are low-resource in many NLP tasks, including Arabic NER.

| Dialects/Domains | Politics | Economics | Finance | History | Law | Science | Health | Agriculture | Art | Sport | Total |
|---|---|---|---|---|---|---|---|---|---|---|---|
| 1 MSA | 12712 | 8010 | 8149 | 8165 | 8083 | 8378 | 8777 | 8882 | 8015 | 9386 | 88557 |
| 2 Syria | 4020 | 4025 | 4043 | 4004 | 4018 | 4139 | 4018 | 4151 | 7648 | 4055 | 44121 |
| 3 Palestine | 4447 | 4511 | 4006 | 4455 | 4021 | 4253 | 4580 | 4002 | 4012 | 4036 | 42323 |
| 4 Lebanon | 4396 | 4205 | 4341 | 4163 | 4173 | 4002 | 4229 | 4085 | 5939 | 4131 | 43664 |
| 5 Saudi Arabia | 4270 | 4165 | 4162 | 4104 | 3996 | 4036 | 4079 | 4079 | 4650 | 4054 | 41595 |
| 6 Oman | 5094 | 4049 | 4484 | 4209 | 5275 | 4123 | 5024 | 4227 | 4023 | 4486 | 44994 |
| 7 Yemen | 5312 | 4375 | 4212 | 4045 | 5479 | 4100 | 4000 | 4011 | 4242 | 4042 | 43818 |
| 8 Iraq | 4000 | 4777 | 4014 | 4003 | 4007 | 4000 | 4012 | 4008 | 4045 | 4002 | 40868 |
| 9 Kuwait | 4148 | 4656 | 5322 | 3998 | 5894 | 4022 | 4518 | 4044 | 4013 | 4073 | 44688 |
| 10 Egypt | 4096 | 4085 | 4306 | 4124 | 4141 | 4196 | 4041 | 4286 | 4095 | 4072 | 41442 |
| 11 Sudan | 4145 | 4508 | 4078 | 4004 | 4138 | 4207 | 4371 | 4158 | 4135 | 4013 | 41757 |
| 12 Libya | 11898 | 4172 | 4041 | 5775 | 5350 | 4307 | 6757 | 16044 | 4151 | 4802 | 67297 |
| 13 Tunisia | 4113 | 4594 | 4019 | 14918 | 4778 | 4070 | 6073 | 4787 | 9159 | 5170 | 61681 |
| 14 Algeria | 3073 | 3030 | 3025 | 3294 | 3050 | 3016 | 3557 | 3059 | 3063 | 3069 | 31236 |
| 15 Morocco | 5378 | 4913 | 5193 | 5019 | 4107 | 4078 | 4887 | 4079 | 8989 | 4268 | 50911 |
| 16 Mauritania | 6991 | 5424 | 4007 | 5492 | 4015 | 4050 | 4620 | 4039 | 4005 | 6147 | 48790 |
| Total | 88093 | 73499 | 71402 | 83772 | 74525 | 68977 | 77543 | 81941 | 84184 | 73806 | 777742 |

Figure 1: *Konooz* statistics, by domain and dialect.

MSA benefits from a large pool of annotated NER resources, while dialects suffer from a lack of such datasets (Khbir et al., 2023). Additionally, existing datasets have mainly focused on a limited number of domains, such as *Wojood* (Jarrar et al., 2022, 2023a) which covers two dialects and five domains. Other resources such as *OntoNotes* (Weischedel et al., 2013b) and *ANERCorp* (Benajiba et al., 2007) focus exclusively on MSA and are limited to the political news domain. The lack of labeled datasets for multiple dialects and domains makes developing and evaluating NER models in cross-domain and cross-dialect more challenging (Mekki et al., 2022; Jia et al., 2019).

*Konooz* is a novel multi-dimensional NER corpus designed for benchmarking NER models across various domains and dialects. To the best of our knowledge, *Konooz* is the first rich corpus that contains 10 domains in 16 different dialects. As shown in Figure 1, each corpus contains about 4k tokens. The MSA corpus contains about 8k tokens. The corpus was manually collected from diverse sources, reflecting the diversity of Arabic dialects. Then, it was manually annotated by a team of 45 people. The annotation process involved labeling tokens with 21 distinct entity types, utilizing both flat and nested NER tags, adhering to the annotation guidelines presented in Jarrar et al. (2022).

We used *Konooz* for benchmarking four Arabic NER models, which revealed low performance across domains and dialects. The variations in the results underscore the urgent need for more diverse Arabic NER datasets. Furthermore, we leveraged *Konooz* to conduct an in-depth analysis of lexical similarity across domains and dialects. We hypothesized that (i) dialects from the same geographic region are expected to exhibit low divergence, and (ii) the named entities from the same country or region in the training data improve model performance. By measuring the divergence between domains and dialects, we uncovered several insightful patterns and correlations. These divergences reveal the linguistic variations that directly impact the performance degradation of the trained models. Efficiently measuring and reducing divergence is crucial for adapting models to the new domain—the topic of domain adaptation (Kashyap et al., 2021).

In short, the main contributions of this paper are:

1. *Konooz*, 160 corpora covering 10 domains and 16 dialects (777k tokens) manually labeled with 21 entity types in flat and nested annotations.

2. Benchmark of four Arabic NER models using *Konooz* in cross-domain and cross-dialect.

3. Insightful lexical similarity analysis that uncovers distinctions and similarities across different domains and dialects.

The paper is organized as: Section 2 reviews related work; Section 3 presents *Konooz*; Section 4 reveals the lexical similarity; Section 5 benchmarks NER models; and we conclude in Section 6.

## 2 Related Work

### 2.1 Arabic NER Datasets

Most available NER corpora cover MSA with limited coverage of domains and dialects. *ANER-Corp* is an Arabic news corpus comprising $150k$ tokens annotated with four entity types (Benajiba et al., 2007). *OntoNotes* 5.0 includes $300k$ MSA tokens annotated with 17 entity types (Weischedel et al., 2013a). *Wojood* is a larger corpus containing $550k$ annotated for both flat and nested entities. *Wojood* uses the 17 entity types used in *OntoNotes*, and introduces four additional types: Occupation (OCC), WEBSITE, UNIT, and Currency (CURR) (Jarrar et al., 2022). *Wojood* was later extended with $Wojood_{fine}$, $Wojood^{Hadath}$ and $Wojood^{Gaza}$. $Wojood_{fine}$ is the same as *Wojood*, but it introduces 30 fine-grained sub-entity types (Liqreina et al., 2023; Jarrar et al., 2023a). $Wojood^{Hadath}$ is the Wojood corpus annotated with event argument relations (Aljabari et al., 2024). $Wojood^{Gaza}$ consists of 60k tokens (Jarrar et al., 2024) focusing on news related to the Israeli War on Gaza across and Nakba NLP (Jarrar et al., 2025). It covers five domains (politics, law, economy, finance, and health) annotated with 51 entity types and subtypes following Wojood guidelines.

There are only a few dialectal NER corpora covering limited number of entity types. Zirikly and Diab (2014) presented an Egyptian NER corpus containing 40k tokens. DarNERCorp is a Moroccan NER corpus comprising 65k tokens annotated with four entity types (Moussa and Mourhir, 2023). NERDz is an Algerian corpus annotated with eight entity types (Touileb, 2022). DzNER is another Algerian corpus (220k) annotated with three entity types (Dahou and Cheragui, 2023).

Several publicly available dialectal corpora exist, such as the *Lisan* corpora (Jarrar et al., 2023c), which covers Iraqi, Libyan, Sudanese, and Yemeni dialects; the $Nabra^{Syrian}$ corpus (Nayouf et al., 2023); and the *Curra+Baladi* corpora for Palestinian and Lebanese dialects (Haff et al., 2022; Jarrar et al., 2017). All of these corpora are fully annotated with morphological tags and lemmas linked with Qabas (Jarrar and Hammouda, 2024) and the Arabic Ontology (Jarrar, 2021). However, among these, only *Curra+Baladi* is annotated with NER tags, as it is part of the Wojood dataset. Other spoken dialectal corpora with transcriptions exist, such as Casablanca (Talafha et al., 2024), but none include NER annotations.

In other languages, BarNER is the first manually annotated NER dataset, comprising 161k tokens sourced from Bavarian Wikipedia articles and tweets, annotated according to a schema adapted from the German CoNLL 2006 guidelines. The dataset includes both coarse-grained and fine-grained entity types (Peng et al., 2024).

### 2.2 Benchmarking NER Models

Previous research has focused on building benchmark datasets and evaluating NER models. Vajjala and Balasubramaniam (2022) challenge the reliance on micro-$F_1$ scores and propose a broader evaluation framework assessing models across entity categories, sources, and genres. Using *OntoNotes* 5.0 and six adversarial test sets, they evaluate Spacy, Stanza, and SparkNLP, revealing $F_1$-score drops of 12%-20% across sources and genres. This highlights that NER models struggle with unseen genres, even with multi-genre training.

## 3 *Konooz* Corpus

*Konooz* was manually collected and covers 16 dialects, including MSA, with each dialect represented across 10 domains.

### 3.1 Corpus Collection Guidelines

We manually collected dialectal threads from various sources—including Facebook, X, YouTube comments, and blogs—ensuring that only public posts and comments from public accounts were included. For MSA, we retrieved articles from specific domains on public media websites, like AlJazeera, AlArabiya, and SkyNewsArabia.

We collected dialectal threads consisting of one or more sentences, each containing more than five words to ensure sufficient context. Additionally, every sentence is manually categorized into a specific domain by analyzing its context and identifying domain-specific keywords. Each sentence should include multiple entity types, such as person names, organizations, events, and more. All sentences were written between 2010 and 2022. To ensure a balanced dataset, each domain within each dialect must include about 4k tokens.

### 3.2 Collection Procedure

Since recruiting native speakers for each dialect is challenging, we hired 40 students—at a rate of 5 USD per hour — to collect the corpus, taking into account the following measures to ensure high-quality collection:

- **Dialect Familiarization**: To help a student become familiar with the necessary vocabulary in a dialect, we asked the student to watch about two hours of content in that dialect.

- **Dialect and Domain Identification**: We identified local TV and radio stations and located their social media channels to target dialect-specific content. With this strategy, we assumed people not native to the target dialect are less likely to comment on local issues outside their country. We also identified local and domain-specific pages, such as the Homs Agriculture Chamber of Syria, NBK Bank in Assiut-Kuwait, Khamsint Ektesad in Egypt, and Ask Software Engineers in Palestine.

- **Dialect Similarity Verification**: After collecting the corpora, we conducted dialect identification tests. We randomly selected 10% of the sentences from a specific dialect and mixed them with similar dialects (e.g., Syrian and Lebanese). Native speakers were then tasked to identify the dialects of these sentences. The results showed an average of 87% accuracy.

- **MSA Divergence Verification**: To ensure our dialectal corpora are not MSA, we used the Arabic Level of Dialectness (ALDi) model (Keleg et al., 2023). It helps to quantify the divergence of sentences from MSA. Sentences scoring below 0.2 were manually reviewed to confirm they were truly MSA, resulting in the removal of 8% of sentences. The ALDi scores for all dialects are shown in Figure 2.

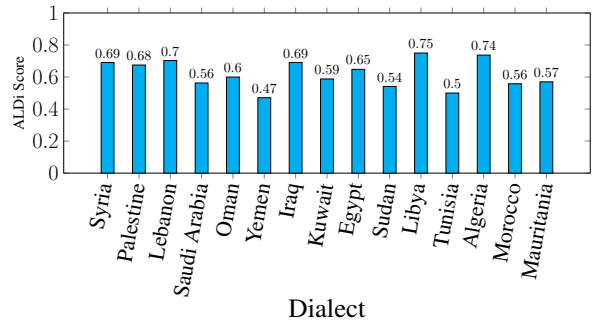

Figure 2: ALDi score for each dialect in *Konooz*.

The collected data was stored in separate Google Sheets for each dialect for manual annotation.

### 3.3 Annotation Methodology

**Phase I** To bootstrap the annotation process, we used the *Wojood* NER model in *SinaTools* (Hammouda et al., 2024) to tokenize and tag the corpus.

**Phase II** We recruited five annotators, each holding a master's degree in linguistics and experience in NER and multilingual annotation projects, at a rate of 8USD/hour. They were introduced to the NER guidelines (Jarrar et al., 2022) to familiarize themselves with 21 entity types (see §A.1). Each annotator was initially assigned $1k$ tokens to annotate, which was reviewed by a NER expert. Once verified, they were assigned to annotate 2-4 dialects over a span of 15 months. Regular group discussions were held to address challenging cases and ambiguities in named entities.

**Phase III** We trained a NER model using the *Wojood* dataset (Jarrar et al., 2022), *Wojood*<sup>Gaza</sup> (Jarrar et al., 2024), and the first version *Konooz* (See the details in §A.2). The trained model is used to annotate the *Konooz* for a second round. Then, the annotators reviewed the results to identify errors. In this round, 1,500 errors were corrected. This process was repeated again, in which we identified 10 changes only, indicating a significant improvement in annotation accuracy.

### 3.4 Annotation Challenges

We faced several annotation challenges. First, annotators often struggle to recognize local and dialect-specific landmarks and place names. They had to search for these mentions in each dialect or consult native speakers for clarification, such as the (جسر الزرقا *ǧsr ālzrqā* /Az-Zarqa Bridge) a GPE in Palestine, (نزل تاب *nzl tāb* ) a FAC in Tunis and (الكولا *ālkwlā* ) a FAC in Lebanon. In contrast, identifying people's names is much more straightforward. Second, certain named entities, such as dates, times, and numbers, vary across dialects, which makes annotation more challenging. For example, the number (اثنين *āṯnyn* /two) in Saudi, Omani, and Yemeni is (جوج *ǧwǧ* ) in Moroccan. The word (الآن *ālʾān* /now) in MSA is (دابا *dābā* ) in Moroccan, (تو *tw* ) in Omani, (للحين *llḥyn* ) in Kuwaiti, (هلا *hlā* ) in Syrian, and (هسا *hsā* ) in Palestinian. To handle such variations, annotators engaged in group discussions and consulted native speakers to ensure accurate labeling.

### 3.5 Inter-Annotator Agreement

To evaluate the consistency of our annotations, we used Cohen's Kappa (Cohen, 1960), a standard metric for inter-annotator agreement (IAA) (Hripcsak and Rothschild, 2005). We randomly selected about 7% (39k tokens) from each domain in each dialect to be annotated by another annotator. We calculated both Kappa and $F_1$ scores. Table 1 demonstrates high agreement in all entity types.

The high IAA can be attributed to several factors. Continuous feedback to annotators from native speakers and periodic discussions ensured consistency during the process. The annotation conducted during Phase III was instrumental in enhancing consistency among annotators. Comparing annotator outputs with model annotations provided valuable insights, especially in cases where the model correctly identified entities missed by annotators. This collaborative and human-in-the-loop approach significantly improved the overall data quality.

| Entity Type | TP | FN | FP | $\mathcal{K}$ | $F_1$-Score |
|---|---|---|---|---|---|
| ORG | 1255 | 8 | 25 | 0.987 | 0.987 |
| DATE | 950 | 17 | 16 | 0.983 | 0.983 |
| WEBSITE | 59 | 1 | 1 | 0.983 | 0.983 |
| OCC | 594 | 6 | 11 | 0.986 | 0.986 |
| CURR | 94 | 0 | 2 | 0.989 | 0.989 |
| PERS | 975 | 4 | 12 | 0.992 | 0.992 |
| LAW | 71 | 0 | 0 | 1 | 1.000 |
| PRODUCT | 213 | 1 | 13 | 0.968 | 0.968 |
| EVENT | 394 | 0 | 12 | 0.985 | 0.985 |
| GPE | 985 | 5 | 12 | 0.991 | 0.991 |
| NORP | 1122 | 18 | 52 | 0.970 | 0.970 |
| UNIT | 77 | 2 | 0 | 0.987 | 0.987 |
| LANGUAGE | 25 | 0 | 0 | 1 | 1.000 |
| TIME | 234 | 4 | 6 | 0.979 | 0.979 |
| MONEY | 170 | 2 | 0 | 0.994 | 0.994 |
| LOC | 236 | 10 | 5 | 0.969 | 0.696 |
| QUANTITY | 150 | 5 | 0 | 0.984 | 0.984 |
| PERCENT | 195 | 0 | 0 | 1 | 1.000 |
| CARDINAL | 521 | 6 | 4 | 0.99 | 0.99 |
| ORDINAL | 283 | 4 | 20 | 0.961 | 0.961 |
| FAC | 123 | 0 | 8 | 0.969 | 0.969 |
| **Overall** | **8726** | **93** | **199** | **0.984** | **0.971** |

Table 1: IAA for each entity type.

### 3.6 *Konooz* Statistics

*Konooz* comprises 31,265 sentences, with an average sentence length of 28.18 words, annotated nested and flat entities, all tagged with 21 coarse-grained tags. Table 2 presents the overall statistics, while Table 11 in §A.5 provides detailed statistics.

| Tag | Flat | Nested | Total |
|---|---|---|---|
| PERS | 9,564 | 652 | 10,216 |
| ORG | 8,512 | 1213 | 9,725 |
| LOC | 1,680 | 235 | 1,915 |
| GPE | 9,947 | 1,969 | 11,916 |
| NORP | 11,583 | 494 | 12,077 |
| CARDINAL | 6,764 | 92 | 6,856 |
| ORDINAL | 4,350 | 344 | 4,694 |
| OCC | 6,270 | 91 | 6,361 |
| FAC | 757 | 28 | 785 |
| PRODUCT | 746 | 10 | 756 |
| EVENT | 1,612 | 66 | 1,678 |
| DATE | 7,526 | 195 | 7,721 |
| TIME | 2,432 | 4 | 2,436 |
| LANGUAGE | 315 | 2 | 317 |
| WEBSITE | 410 | 4 | 414 |
| LAW | 369 | 3 | 372 |
| PERCENT | 810 | 4 | 814 |
| QUANTITY | 827 | 13 | 840 |
| UNIT | 125 | 773 | 898 |
| MONEY | 1,495 | 67 | 1,562 |
| CURR | 974 | 1128 | 2,102 |
| Total | 77,068 | 7,387 | 84,455 |

Table 2: Statistics about NER annotations in *Konooz*.

## 4 Lexical Similarity Analysis

The performance of trained models on out-of-distribution data is heavily influenced by data distribution divergence. Efficiently measuring and minimizing this divergence is crucial for effective domain adaptation (Ben-David et al., 2010). Lexical similarity measures vocabulary overlap between different data distributions (i.e., domains and dialects). It is used to identify source domains or dialects most aligned with the target distribution to enhance domain adaptation (Dai et al., 2019). Domains and dialects that have the same underlying distribution often exhibit high lexical similarity (Pogrebnyakov and Shaghaghian, 2021). We hypothesize that dialects from the same geographic region are expected to exhibit low divergence. To verify this, different metrics are utilized to measure the overlap between dialects and domains.

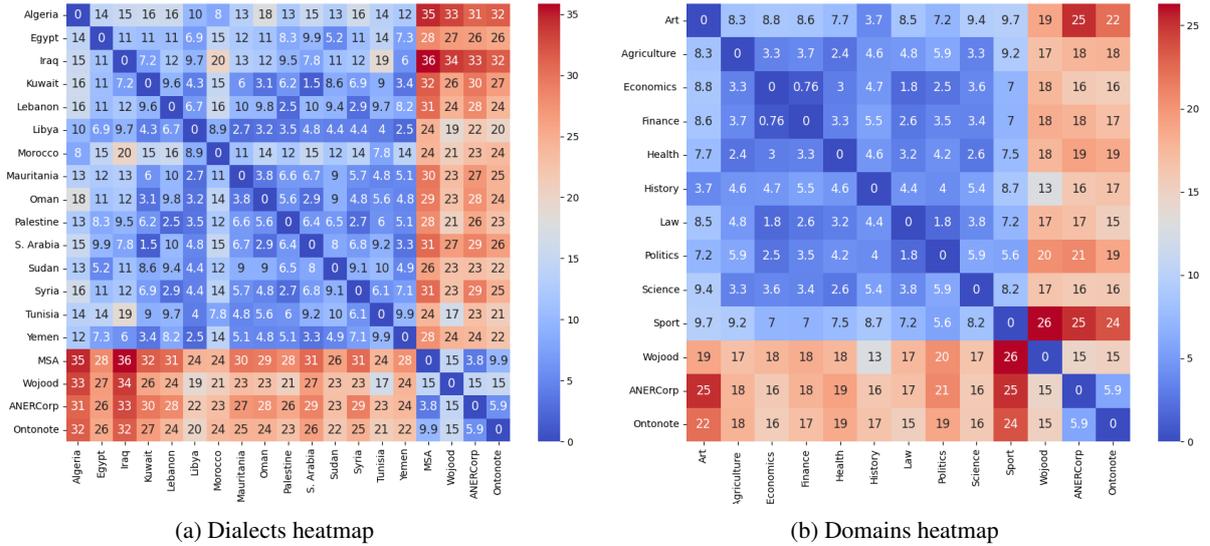

(a) Dialects heatmap  (b) Domains heatmap

Figure 3: Heatmaps of the MMD distances among *Konooz* dialects and domains using linear kernel and between *Konooz*, *Wojood*, *ANERCorp* and *OntoNotes* (last three rows).

Jaccard similarity coefficient and cosine similarity metrics failed to differentiate between dialects and domains, as shown in §A.5, and according to Kashyap et al. (2021) Maximum Mean Discrepancy (MMD) (Gretton et al., 2008) outperforms traditional similarity metrics like cosine similarity and KL divergence, making it a more reliable indicator of performance shifts across domains. We used AraBERTv2 (Antoun et al., 2020) sentence representations as the bases for MMD and revealed lexical similarity variations from 1.1 to 13 across domains and 1.5 to 36 across dialects, where higher values indicate dissimilarity (see Figure 3).

**Dialect Lexical Similarity** Figure 4 is a t-SNE plot generated using the AraBERTv2 sentence representations to visualize the overlap among dialects, including MSA. One can clearly see MSA (red cluster) is notably distinct from the other dialects. Some dialects form compact clusters, such as Moroccan (fuchsia color), Algerian (light green) and Iraqi (yellow), while others show more overlap. Figure 3a quantifies the distances among the clusters in Figure 4. The results reveal significant variations between the dialects. The highest scores are observed between MSA and other dialects, with the highest MMD of 36 recorded between MSA and Iraqi. This indicates significant differences in vocabulary and contextual usage between the MSA and the other dialects. Conversely, the lowest MMD score of 1.5 is observed between the Kuwaiti and Saudi dialects. Such low divergence indicates that data comes from closely related dialects and reflects their close linguistic and contextual similarities, as well as their shared cultural and geographic ties within the Gulf region.

The Moroccan dialect demonstrates the highest level of dissimilarity, aligning with expectations due to its pronounced divergence from other Arabic dialects. Its distinct phonetic, lexical, and syntactic features differentiate it significantly, particularly from Gulf and Levantine dialects. In datasets with high divergence, models are more likely to generate lower confidence scores or misclassify entities, highlighting the need for dialect-specific adaptations to improve performance.

Figure 3a has two distinct clusters. The 15 dialects (excluding MSA) cluster in the top left of the figure and a smaller cluster containing MSA,

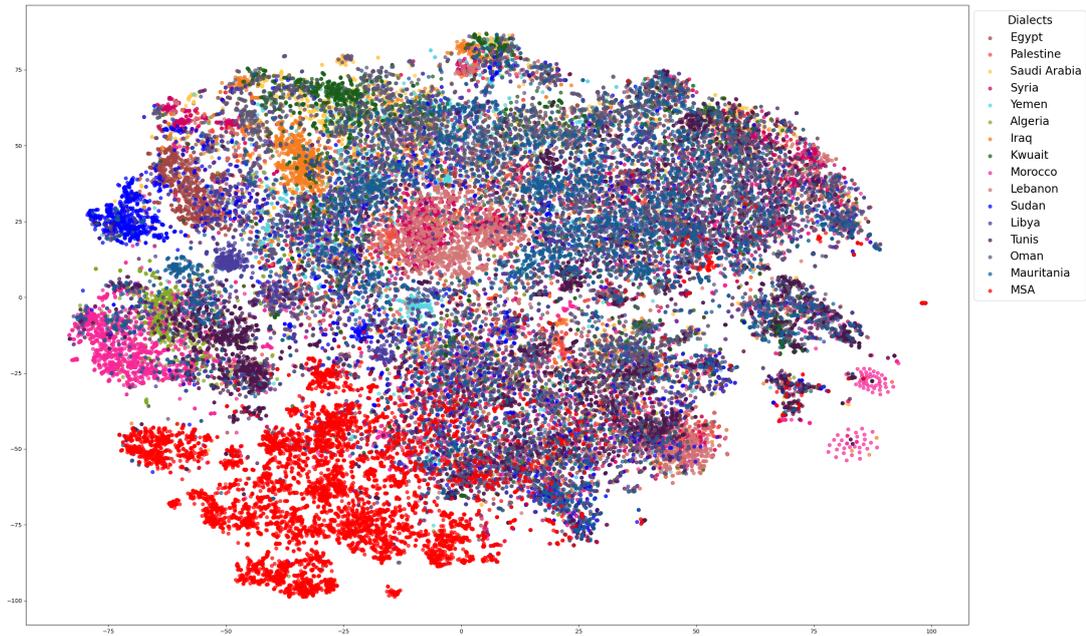

Figure 4: t-SNE visualization of high-dimensional feature embeddings of *Konooz* dialects.

*Wojood*, *ANERCorp* and *OntoNotes* datasets (the MMD values in the smaller cluster are computed between MSA and the entirety of the three training datasets). The most important observation is the inclusion of MSA and the three training datasets in one cluster, which demonstrates the heavy representation of MSA in the NER datasets. Wojood includes Palestinian and Lebanese data; however, this subset represents only 12% of the corpus. As a result, the lexical distances for these dialects are not the lowest in Figure 3a. One possible explanation is limited overlap between the training and test sets in terms of domain and named entities. Regarding to some dialects such as Tunisia, which shows the lowest lexical distance despite not being explicitly represented in *Wojood*, our hypothesis is that shared named entities—especially in formal contexts may contribute to this reduced surface-level lexical distance across dialects. In these cases, entity classification becomes more challenging as the model encounters unfamiliar linguistic structures and patterns. Effective domain adaptation techniques are needed to bridge the gap between dialects and MSA.

Finally, we measured the MMD between MSA and dialectal data across domains to assess genre influence. Figure 5 shows that intra-domain MMDs (diagonal) are high, indicating significant divergence even within the same domain. Some inter-domain MMDs (off-diagonal) show even greater divergence, highlighting the impact of domain differences. As noted earlier, MSA data is sourced from online news articles, which follow a rigid structure and have clear domain distinctions. In contrast, dialectal data is collected from social media platforms, where it is written informally and may lack domain-specific vocabulary.

**Domain Lexical Similarity** Figure 3b highlights the significant lexical differences among domains. The greatest dissimilarity is observed between the Art and Science and Art and Sport, with a value of 13. This can be attributed to specific and unique topics discussed in these domains. Art corpora often reference creative works, artists, and cultural institutions, whereas Science corpora emphasize research, scientific disciplines, and technical terminology. The difference in entity types and contexts leads to a small semantic overlap in vocabulary. In contrast, the highest similarity is found between Finance and Economics, with a value of 1.1. This indicates that Finance and Economics share the strongest overlap, likely due to common topics like markets, investments, and risk.

We also measured the MMD between the do-mains of *Konooz* and the entirety of *Wojood*, *ANERCorp*, and *OntoNotes*. As shown in the lower right part of Figure 3b, the MMD is higher compared to that between other domains, indicating greater divergence. This aligns with the significant performance degradation discussed in Section 5.2.

Measuring divergence highlights the level of effort needed to adapt a model trained on one domain to perform effectively in another. Our analysis aims to provide an indicator of lexical variation for cross-domain transfer learning rather than a definitive measurement of language differences, which would require comprehensive coverage of all dialects.

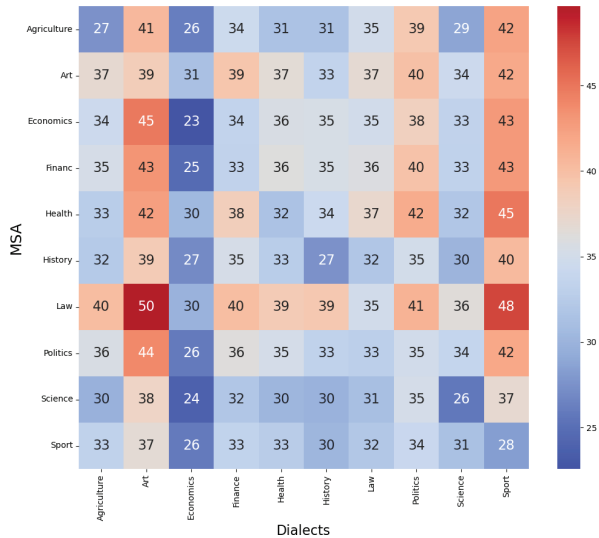

Figure 5: Heatmap showing the genre differences between the MSA and dialectal data.

## 5 NER Benchmarks using *Konooz*

This section benchmarks four state-of-the-art Arabic NER models in cross-domain and cross-dialect scenarios using *Konooz*. First, we conducted in-distribution evaluations by training and evaluating four models using their testing datasets (*Wojood$_{Nested}$*, *Wojood$_{Flat}$*, *OntoNotes* and *ANERCorp*). Second, we benchmarked these models against the 10 domains (out-of-domain evaluation) and the 16 dialects of *Konooz* (out-of-dialect evaluation). Since the four NER models follow different annotation guidelines, we mapped *Konooz* entity types to align with each training data accordingly. The mapping and the benchmarking preparations are provided in Appendix §A.6. We named the models based on their corresponding trained dataset.

### 5.1 Cross-Dialect Benchmarking

We benchmarked the four NER models across the 16 dialects of *Konooz*. As shown in Table 3, all models demonstrated a significant drop in performance when evaluated on *Konooz* (out-of-dialect). We observed performance degradation in cross-dialect evaluation ranging from 25% to 30%. There are multiple patterns can be observed. First, three dialects have demonstrated higher performance across all models, which are MSA, Lebanese, and Egyptian. *Wojood$_{Nested}$* and *Wojood$_{Flat}$* achieved the highest performance on *Konooz* MSA - about 20% performance degradation. This is expected since the majority of Wojood training data consists of MSA collected in the same period (2010-2024). The *OntoNotes* performed better on Lebanese, which could be because of the OntoNotes 2,099 named entity mentions related to Lebanon. Similarly, the ANERCorp model performed better on Egyptian because ANERCorp covers more news related to Egypt. Based on insights from Figure 3a, it is expected that MSA is expected to be one of the top performing dialects as it is closely clustered with the training data of *Wojood*, *OntoNotes*, and *ANERCorp*.

Second, aside from the dialects we discussed above, we see significant drop in performance for all the other dialects across all NER models because most MMD scores between the training datasets and *Konooz* dialects are significantly high reaching 20s to mid 30s. The dialectal differences limit the models' ability to recognize and adapt to unfamiliar linguistic patterns, including variations in vocabulary, syntax, and morphological structures. As shown in Figures 3a and 4, the MSA embeddings (red clusters) are scattered away from those of other dialects, indicating poor alignment in the feature space. This misalignment illustrates the models' struggle to generalize across dialects without robust domain adaptation techniques to bridge the gap between different domains.

Third, differences in annotation guidelines influence performance. Since *Konooz* follows the *Wojood* guideline (Jarrar et al., 2022), this may partly explain why *Wojood$_{Nested}$* and *Wojood$_{Flat}$* performed better. Table 5 presents the micro $F_1$-score per entity for all four models. The entity type with the highest confidence across all models was GPE, likely because geopolitical entities (e.g., countries and cities) remain consistent across domains and dialects. Additionally, PERS and PERCENT achieved high scores, as person names and percentage values are easier to identify. Nested annotations add another layer of complexity by capturing hierarchical relationships, such as a "university" (organization) within a "city" (location), which can vary across dialects. On the other hand, the lowest confidence for EVENT, LAW, PRODUCT, and WEBSITE, is likely due to their higher domain specificity.

| NER Models | In-dialect Performance | MSA | Syria | Palestine | Lebanon | S. Arabia | Oman | Yemen | Iraq | Kuwait | Egypt | Sudan | Libya | Tunisia | Algeria | Morocco | Mauritania | Average | Performance Degradation |
|---|---|---|---|---|---|---|---|---|---|---|---|---|---|---|---|---|---|---|---|
| $Wojood_{Nested}$ (21 tags) | 92% | **73%** | **55%** | **65%** | **68%** | **61%** | **61%** | **66%** | **67%** | **65%** | **68%** | **62%** | **64%** | **63%** | **65%** | **63%** | **55%** | **64%** | 28% |
| $Wojood_{Flat}$ (21 tags) | 90% | 71% | 49% | 59% | 62% | 60% | 56% | 61% | 62% | 61% | 61% | 60% | 60% | 58% | 55% | 59% | 52% | 59% | 30% |
| *OntoNotes* (18 tag) | 68% | 45% | 37% | 41% | 51% | 42% | 42% | 41% | 40% | 45% | 45% | 44% | 39% | 38% | 41% | 46% | 29% | 42% | 26% |
| *ANERCorp* (3 tags) | 84% | 58% | 46% | 45% | 64% | 54% | 43% | 58% | 54% | 54% | 66% | 55% | 48% | 41% | 36% | 38% | 37% | 59% | 25% |

Table 3: Micro-$F_1$, NER models trained on datasets (in-dialect) and benchmarked on *Konooz* dialects (out-of-dialect). $Wojood_{Nested}$ is the model trained on the nested version of Wojood dataset, $Wojood_{Flat}$ is trained on the flat version of Wojood, *ANERCorp* is trained on the ANERCorp dataset, and *OntoNotes* is trained on the OntoNotes dataset.

| NER Models | In-domain Performance | Politics | Economics | Finance | History | Law | Science | Health | Agriculture | Art | Sport | Average | Performance Degradation |
|---|---|---|---|---|---|---|---|---|---|---|---|---|---|
| $Wojood_{Nested}$ (21 tags) | **92%** | **68%** | **66%** | **66%** | **68%** | **65%** | 63% | **67%** | **62%** | **63%** | **59%** | **63%** | 29% |
| $Wojood_{Flat}$ (21 tags) | 90% | 64% | 60% | 58% | 66% | 59% | **64%** | 65% | 60% | 60% | 54% | 60% | 30% |
| *OntoNotes* (18 tags) | 68% | 42% | 35% | 34% | 39% | 36% | 35% | 39% | 35% | 43% | 43% | 37% | 31% |
| *ANERCorp* (3 tags) | 84% | 48% | 46% | 40% | 53% | 41% | 41% | 40% | 48% | 54% | 48% | 46% | 38% |

Table 4: Micro-$F_1$, NER models trained on datasets (in-domain) and benchmarked on *Konooz* domains (out-domain).

| Tag | $Wojood_{Nested}$ | $Wojood_{Flat}$ | OntoNotes | ANERCorp |
|---|---|---|---|---|
| CARDINAL | 0.5552 | 0.5605 | 0.1754 | - |
| CURR | 0.7853 | 0.0089 | - | - |
| DATE | 0.6165 | 0.5646 | 0.3508 | - |
| EVENT | 0.3531 | 0.3160 | 0.1835 | - |
| FAC | 0.4378 | 0.4641 | 0.2775 | - |
| GPE | 0.8298 | 0.7775 | 0.5422 | - |
| LANGUAGE | 0.4811 | 0.4991 | 0.0000 | - |
| LAW | 0.3478 | 0.1809 | 0.1044 | - |
| LOC | 0.4798 | 0.5313 | 0.1296 | 0.6403 |
| MONEY | 0.6442 | 0.6274 | 0.4517 | - |
| NORP | 0.5923 | 0.5774 | 0.2450 | - |
| OCC | 0.6293 | 0.5790 | - | - |
| ORDINAL | 0.6176 | 0.5922 | 0.4760 | - |
| ORG | 0.5539 | 0.5362 | 0.4033 | 0.2769 |
| PERCENT | 0.6752 | 0.7633 | 0.0811 | - |
| PERS | 0.7271 | 0.7317 | 0.5365 | 0.5503 |
| PRODUCT | 0.0143 | 0.0184 | 0.0000 | - |
| QUANTITY | 0.2577 | 0.5770 | 0.3776 | - |
| TIME | 0.4242 | 0.4446 | 0.0739 | - |
| UNIT | 0.3780 | 0.0000 | - | - |
| WEBSITE | 0.3056 | 0.2774 | - | - |
| Micro $F_1$ | 0.6316 | 0.6065 | 0.3702 | 0.4600 |
| Macro $F_1$ | 0.4458 | 0.4585 | 0.2449 | 0.3892 |

Table 5: Micro $F_1$-score per entity in *Konooz*, across all dialects and domains.

### 5.2 Cross-Domain Benchmarking

We benchmarked state-of-the-art Arabic NER models across the 10 domains of *Konooz*. The results are summarized in Table 4. While $Wojood_{Nested}$ achieved the highest performance (63%), all models showed significant performance degradation in out-of-domain evaluations, with drops ranging from 29% to 38%. $Wojood_{Nested}$ and $Wojood_{Flat}$ excelled in the history domain, which is the domain with the lowest MMD score to *Wojood* as shown in Figure 3b. This is not surprising as the Wojood training data was sourced from Awraq, the Birzeit University digital Palestinian archive, which covers modern history and cultural heritage.

Similar to the dialects, the performance degradation across the four models can be attributed to domain shift. As shown in Figure 3b, the MMD between *Konooz* and the training datasets is significant. This degradation is driven by significant statistical differences between the training and out-of-domain data, explaining the performance drop across all four models. These models lack domain-invariant features, which severely limits their ability to generalize beyond the specific characteristics of the out-of-domain data and ultimately hinders their performance. Additionally, as shown in Figure 11 in Appendix 7, the *Konooz* domains are scattered inconsistently, emphasizing the significant domain shift between *Konooz* domains.

### 5.3 Discussion

Although lexical similarity provides some indication of model performance, it cannot be directly leveraged to enhance that performance. For example, *Wojood* and the Iraqi corpus are lexically dissimilar (in Figure 3a), however, *Wojood* still achieved relatively-good results in recognizing named entities in the Iraqi corpus (Table 3). After reviewing all cases manually, we found that many of the names of people and geographical places are often shared and recognizable across both MSA/Wojood and Iraqi. This suggests that high lexical dissimilarity does not necessarily lead to poor NER performance. For example, many phrases related to Iraq (Iraq العراق *ālʕrāq* , Iraqi العراقي *ālʕrāqy* , اطفال العراق *āṭfāl ālʕrāq* Iraqi children) appear over 700 times in *Wojood*, which contributes to enhancing the performance of the model. Moreover, *Wojood* includes Palestinian and Lebanese data; however, this subset represents only 12% of the corpus. As a result, the lexical distances for these dialects are not the lowest in Figure 3a.

Furthermore, the improved performance of *OntoNotes* on Lebanese and *ANERCorp* on Egyptian may be better explained by entity coverage rather than overall lexical similarity. *ANERCorp* achieved the best result on Egyptian dialect with an $F_1$ of $66\%$, while *OntoNotes* obtained the best result on Lebanese, with $F_1$ of $51\%$. The *ANERCorp* and *OntoNotes* models are both trained primarily on MSA news. *ANERCorp* performs better on Egyptian, likely because it is trained on data with Egyptian content, particularly from news sources. Similarly, *OntoNotes* performs better on Lebanese, as it includes $2,099$ named entities related to Lebanon. As noted, Figure 3a does not indicate low lexical distances for these pairs, suggesting that general language overlap is not the main factor. This supports our hypothesis that having named entities from the same country or region in the training data improves model performance, regardless of dialectal or lexical proximity.

## 6 Conclusion

We presented *Konooz*, a novel multi-dimensional corpus covering 16 dialects across 10 domains, resulting in 160 distinct corpora (777k tokens) annotated with 21 entity types. Our in-depth lexical similarity analysis reveals both distinctions and similarities across different domains and dialects. Our benchmarking of four Arabic NER models in cross-domain and cross-dialect scenarios highlighted the challenges inherent to Arabic dialects, revealing that models trained on MSA struggle to generalize effectively to dialectal Arabic.

## Limitations

Although various measures were implemented to ensure the consistency and quality of the annotations, the annotation process was carried out by non-native speakers. Furthermore, AraBERTv2 embeddings were utilized to calculate MMD, which may influence the results due to model-specific embedding characteristics. Comparing the results obtained using other models, such as CamelBERT, ArBERT, and LLMs, would provide a broader perspective on performance and help identify the most effective embeddings for capturing cross-domain and cross-dialect variations. Moreover, the WojoodNER tokenizer is based on the AraBERT tokenizer, which is primarily trained on MSA text. This affects its ability to tokenize dialectal Arabic effectively, as it may struggle with out-of-vocabulary words. Additionally, our evaluations are influenced by the mappings between the three NER tagsets (*Wojood*, *OntoNotes* and *ANERCorp* ) and the *Konooz* annotation guidelines.

## 7 Ethical Considerations

*Konooz* was collected from publicly available sources, including Facebook, X, YouTube, and blogs. However, we manually reviewed the content to avoid including private or sensitive information.

## Acknowledgments

We would like to acknowledge the partial support that we received from Birzeit University for the project (No. 255176). The authors also thank the annotation team for their efforts— Shaymma Hamayal, Lina Duaibes, Haneen Liqreina, and Rawa Assi. Thanks to Tymaa Hammouda and Nabila Taha for their technical support throughout the project. We also thank the data collection team (Saja Hamayal, Manar Shawhneh, Ayat Halaseh, Hiba Jaouni, Zainab Jaradat, and many others) and the native speakers who participated in the dialect similarity verification task. We would like to thank Palestine Technical University—Kadoorie for providing support.

## A  Appendix

### A.1  Named Entity classes

The *Konooz* dataset was annotated with 21 entity types, as summarized in Table 6, which provides brief descriptions and examples for each type.

### A.2  Human-In-The-Loop NER Model

The purpose of the Human-In-The-Loop NER Mode is to assess *Konooz* annotation consistency and quality. For this model, we utilized the *Wojood* dataset (Jarrar et al., 2022), *Wojood$^{Gaza}$* provided in Subtask 3 of the WojoodNER 2024 shared task (Jarrar et al., 2024), and the initial version of the annotated *Konooz* dataset.

The Wojood dataset is divided into training (385k tokens, 70%), validation (55k tokens, 10%), and test (110k tokens, 20%) sets. The *Wojood$^{Gaza}$* dataset includes 60k tokens, collected and annotated specifically for the shared task. Additionally, the *Konooz* dataset contains 553, 844 tokens. All datasets adhere to the CoNLL format. For training, we used the training splits from *Wojood*, *Wojood$^{Gaza}$*, and *Konooz*, while the validation and test splits from *Wojood* were used for model evaluation. The total size of training data is 1, 265, 144 tokens.

We fine-tuned AraBERTv2 (Antoun et al., 2020) on nested NER tasks with a learning rate of $\eta = 1e^{-5}$, a batch size of 8, and a maximum of 50 epochs, employing early stopping if validation performance did not improve for five consecutive epochs. The model generally converged around epoch 39. The source code is publicly available on GitHub[1].

---

[1] https://github.com/SinaLab/ArabicNER

| Tag | Short Description |
|---|---|
| PERS | People names, - first, middle, last, and nicknames. Titles are not included except Prophets, kings. |
| NORP | Group of people. |
| OCC | Occupation or professional title. |
| ORG | Legal or social bodies - institutions, companies, academies, teams, parties, armies, governments. |
| GPF | Geopolitical entities like countries, cities, and states. |
| LOC | Geographical locations (Non-GPE), rivers, seas, mountains, and other geographical regions. |
| FAC | Name of a specific place, like roads, cafes, buildings, airports, and gates. |
| PRODUCT | Vehicles, weapons, foods, etc. |
| EVENT | Name of an event of general interest, - battles, wars, sports events, demonstrations, disasters, conferences, national/religious days. Place and date are included in the event name. |
| DATE | Reference to specific or relative dates including day, era, duration, month, and year. Characters to separate date components are part of the tag. |
| TIME | Specific or relative times which is less than a day, including day times like evening and night. |
| LANGUAGE | Named human language or named dialect. |
| WEBSITE | Any named website or specific URL. |
| LAW | Reference to legal text like a constitution, acts, contracts, or agreements. |
| CARDINAL | Numerals written in digits or words. |
| ORDINAL | Any ordinal number, digits or words, that does not refer to a quantity. |
| PERCENT | A word or a symbol referring to a percent. |
| QUANTITY | Any value measured by standard units, except dates, times, and money. |
| UNIT | A word or symbol referring to a unit. |
| MONEY | Absolute monetary value, including currency names. |
| CURR | Any name or symbol referring to currency. |

Table 6: *Konooz* entity types and their description.

## A.3 Maximum Mean Discripancy

MMD compares the distributions of two datasets by first projecting the data into Reproducing Kernel Hilbert Space (RKHS) and then computes the mean distance between two distributions in Hilbert Space. Formally, the MMD is defined as follows:

$$MMD(X,Y) = ||\frac{1}{m}\sum \phi(x_i) - \frac{1}{n}\sum \phi(y_i)||_H \quad (1)$$

where $X$ and $Y$ are two probability distributions between two different domains or dialects, $x_i$ and $y_i$ are the CLS token representation returned by the transformer, and $\phi : X \to H$ is the nonlionear projection to feature representation in RKHS. We experimented with two different kernels, linear and polynomial.

## A.4 Inter-Annotator Agreement

Table 7 presents the IAA at the dialect level in *Konooz*.

| Dialect | TP | FN | FP | $\mathcal{K}$ | $F_1$-Score |
|---|---|---|---|---|---|
| MSA | 6948 | 64 | 61 | 0.9835 | 0.9825 |
| Syria | 3478 | 6 | 8 | 0.9952 | 0.9948 |
| Palestine | 2772 | 14 | 15 | 0.9687 | 0.9677 |
| Lebanon | 3445 | 2 | 2 | 0.9953 | 0.9953 |
| S.Arabia | 3814 | 35 | 41 | 0.9611 | 0.9594 |
| Oman | 3408 | 9 | 9 | 0.9628 | 0.9626 |
| Yemen | 55080 | 249 | 264 | 0.9848 | 0.9857 |
| Iraq | 3458 | 30 | 37 | 0.9192 | 0.9652 |
| Kuwait | 587 | 0 | 1 | 0.9979 | 0.9978 |
| Egypt | 3282 | 4 | 4 | 0.9511 | 0.9960 |
| Sudan | 3421 | 20 | 21 | 0.9814 | 0.9802 |
| Libya | 3426 | 7 | 8 | 0.9928 | 0.9927 |
| Tunisia | 3415 | 23 | 22 | 0.9821 | 0.9817 |
| Algeria | 2874 | 3 | 3 | 0.9979 | 0.9970 |
| Morocco | 3921 | 0 | 0 | 1 | 1 |
| Mauritania | 3413 | 20 | 18 | 0.9071 | 0.9066 |

Table 7: Overall IAA for each dialect in *Konooz*

## A.5 *Konooz* Lexical Similarity and Statistics

This section covers the MMD analysis between dialects and domains using a polynomial kernel, along with word-level lexical similarity calculated using the Jaccard coefficient and cosine similarity. It also includes a t-SNE plot visualizing domain distributions and concludes with a detailed summary of the number of entities within each domain and dialect. The most basic metrics to measure similarity are the Jaccard similarity coefficient and cosine similarity. The Jaccard coefficient similarity is computed using unique words for each dialect and domain, while cosine similarity is measured based on sentence embeddings. However, we found that these methods did not differentiate between dialects or domains. For example, Figures 9 and 10 show that all dialects are homogeneous, exhibiting similar level of overlap (we observed same behavior for domains).

| Tag | Politics | Economics | Finance | History | Law | Science | Health | Agriculture | Art | Sport | Total |
|---|---|---|---|---|---|---|---|---|---|---|---|
| PERS | 1094 / 101 | 510 / 43 | 335 / 29 | 1822 / 237 | 588 / 62 | 427 / 24 | 274 / 30 | 186 / 22 | 2863 / 58 | 1465 / 46 | 9564 / 652 |
| ORG | 1354 / 165 | 1001 / 120 | 1141 / 110 | 467 / 63 | 902 / 176 | 857 / 138 | 509 / 148 | 164 / 16 | 261 / 46 | 1856 / 231 | 8512 / 1213 |
| LOC | 254 / 18 | 160 / 9 | 68 / 11 | 542 / 48 | 71 / 8 | 61 / 7 | 39 / 11 | 256 / 13 | 114 / 12 | 115 / 98 | 1680 / 235 |
| GPE | 1524 / 288 | 1214 / 140 | 776 / 240 | 2435 / 304 | 554 / 104 | 529 / 216 | 623 / 124 | 996 / 68 | 652 / 116 | 644 / 369 | 9947 / 1969 |
| NORP | 1739 / 63 | 1002 / 29 | 678 / 14 | 2930 / 101 | 1155 / 87 | 717 / 26 | 892 / 35 | 383 / 9 | 991 / 18 | 1096 / 112 | 11583 / 494 |
| CARDINAL | 519 / 7 | 856 / 23 | 932 / 5 | 561 / 14 | 702 / 12 | 568 / 2 | 590 / 4 | 757 / 12 | 472 / 3 | 807 / 10 | 6764 / 92 |
| ORDINAL | 337 / 20 | 312 / 22 | 296 / 20 | 455 / 58 | 393 / 74 | 536 / 37 | 425 / 20 | 488 / 30 | 518 / 23 | 590 / 40 | 4350 / 344 |
| OCC | 696 / 12 | 501 / 4 | 370 / 2 | 520 / 18 | 775 / 30 | 554 / 4 | 821 / 7 | 320 / 1 | 777 / 5 | 936 / 8 | 6270 / 91 |
| FAC | 65 / 5 | 51 / 1 | 27 / 0 | 270 / 10 | 37 / 0 | 22 / 3 | 70 / 1 | 59 / 0 | 80 / 4 | 76 / 4 | 757 / 28 |
| PRODUCT | 25 / 2 | 35 / 1 | 148 / 1 | 26 / 0 | 14 / 0 | 267 / 1 | 53 / 0 | 35 / 0 | 117 / 4 | 26 / 1 | 746 / 10 |
| EVENT | 251 / 11 | 121 / 3 | 64 / 3 | 285 / 12 | 143 / 2 | 62 / 1 | 40 / 3 | 27 / 1 | 163 / 9 | 456 / 21 | 1612 / 66 |
| DATE | 471 / 41 | 871 / 20 | 841 / 16 | 1128 / 29 | 605 / 18 | 581 / 10 | 666 / 14 | 999 / 2 | 820 / 13 | 544 / 32 | 7526 / 195 |
| TIME | 177 / 0 | 214 / 1 | 195 / 0 | 443 / 2 | 166 / 1 | 206 / 0 | 305 / 0 | 254 / 0 | 240 / 0 | 232 / 0 | 2432 / 4 |
| LANGUAGE | 8 / 0 | 7 / 0 | 10 / 0 | 104 / 2 | 18 / 0 | 102 / 0 | 5 / 0 | 4 / 0 | 48 / 0 | 9 / 0 | 315 / 2 |
| WEBSITE | 31 / 0 | 34 / 0 | 60 / 1 | 17 / 0 | 22 / 1 | 169 / 2 | 15 / 0 | 14 / 0 | 40 / 0 | 8 / 0 | 410 / 4 |
| LAW | 10 / 0 | 29 / 0 | 21 / 1 | 7 / 0 | 277 / 2 | 3 / 0 | 9 / 0 | 8 / 0 | 2 / 0 | 3 / 0 | 369 / 3 |
| PERCENT | 33 / 0 | 241 / 3 | 203 / 0 | 14 / 0 | 51 / 0 | 42 / 0 | 62 / 0 | 129 / 1 | 23 / 0 | 12 / 0 | 810 / 4 |
| QUANTITY | 20 / 0 | 112 / 3 | 19 / 1 | 30 / 0 | 5 / 0 | 49 / 4 | 43 / 0 | 526 / 3 | 8 / 0 | 15 / 2 | 827 / 13 |
| UNIT | 2 / 20 | 62 / 107 | 4 / 18 | 6 / 29 | 3 / 5 | 4 / 48 | 5 / 44 | 35 / 478 | 4 / 8 | 0 / 16 | 125 / 773 |
| MONEY | 27 / 5 | 496 / 19 | 460 / 30 | 19 / 2 | 88 / 4 | 54 / 0 | 39 / 1 | 234 / 3 | 45 / 0 | 33 / 3 | 1495 / 67 |
| CURR | 27 / 26 | 396 / 352 | 360 / 407 | 16 / 10 | 21 / 79 | 16 / 40 | 9 / 38 | 87 / 129 | 37 / 21 | 5 / 26 | 974 / 1128 |
| Total | 8664 / 784 | 8225 / 900 | 7008 / 909 | 12097 / 939 | 6590 / 665 | 5826 / 563 | 5494 / 480 | 5961 / 788 | 8275 / 340 | 8928 / 1019 | 77068 / 7387 |

Table 8: Total number of entities in each dialect in flat/nested.

| Tag | Politics | Economics | Finance | History | Law | Science | Health | Agriculture | Art | Sport | Total |
|---|---|---|---|---|---|---|---|---|---|---|---|
| PERS | 1094 / 101 | 510 / 43 | 335 / 29 | 1822 / 237 | 588 / 62 | 427 / 24 | 274 / 30 | 186 / 22 | 2863 / 58 | 1465 / 46 | 9564 / 652 |
| ORG | 1354 / 165 | 1001 / 120 | 1141 / 110 | 467 / 63 | 902 / 176 | 857 / 138 | 509 / 148 | 164 / 16 | 261 / 46 | 1856 / 231 | 8512 / 1213 |
| LOC | 254 / 18 | 160 / 9 | 68 / 11 | 542 / 48 | 71 / 8 | 61 / 7 | 39 / 11 | 256 / 13 | 114 / 12 | 115 / 98 | 1680 / 235 |
| GPE | 1524 / 288 | 1214 / 140 | 776 / 240 | 2435 / 304 | 554 / 104 | 529 / 216 | 623 / 124 | 996 / 68 | 652 / 116 | 644 / 369 | 9947 / 1969 |
| NORP | 1739 / 63 | 1002 / 29 | 678 / 14 | 2930 / 101 | 1155 / 87 | 717 / 26 | 892 / 35 | 383 / 9 | 991 / 18 | 1096 / 112 | 11583 / 494 |
| CARDINAL | 519 / 7 | 856 / 23 | 932 / 5 | 561 / 14 | 702 / 12 | 568 / 2 | 590 / 4 | 757 / 12 | 472 / 3 | 807 / 10 | 6764 / 92 |
| ORDINAL | 337 / 20 | 312 / 22 | 296 / 20 | 455 / 58 | 393 / 74 | 536 / 37 | 425 / 20 | 488 / 30 | 518 / 23 | 590 / 40 | 4350 / 344 |
| OCC | 696 / 12 | 501 / 4 | 370 / 2 | 520 / 18 | 775 / 30 | 554 / 4 | 821 / 7 | 320 / 1 | 777 / 5 | 936 / 8 | 6270 / 91 |
| FAC | 65 / 5 | 51 / 1 | 27 / 0 | 270 / 10 | 37 / 0 | 22 / 3 | 70 / 1 | 59 / 0 | 80 / 4 | 76 / 4 | 757 / 28 |
| PRODUCT | 25 / 2 | 35 / 1 | 148 / 1 | 26 / 0 | 14 / 0 | 267 / 1 | 53 / 0 | 35 / 0 | 117 / 4 | 26 / 1 | 746 / 10 |
| EVENT | 251 / 11 | 121 / 3 | 64 / 3 | 285 / 12 | 143 / 2 | 62 / 1 | 40 / 3 | 27 / 1 | 163 / 9 | 456 / 21 | 1612 / 66 |
| DATE | 471 / 41 | 871 / 20 | 841 / 16 | 1128 / 29 | 605 / 18 | 581 / 10 | 666 / 14 | 999 / 2 | 820 / 13 | 544 / 32 | 7526 / 195 |
| TIME | 177 / 0 | 214 / 1 | 195 / 0 | 443 / 2 | 166 / 1 | 206 / 0 | 305 / 0 | 254 / 0 | 240 / 0 | 232 / 0 | 2432 / 4 |
| LANGUAGE | 8 / 0 | 7 / 0 | 10 / 0 | 104 / 2 | 18 / 0 | 102 / 0 | 5 / 0 | 4 / 0 | 48 / 0 | 9 / 0 | 315 / 2 |
| WEBSITE | 31 / 0 | 34 / 0 | 60 / 1 | 17 / 0 | 22 / 1 | 169 / 2 | 15 / 0 | 14 / 0 | 40 / 0 | 8 / 0 | 410 / 4 |
| LAW | 10 / 0 | 29 / 0 | 21 / 1 | 7 / 0 | 277 / 2 | 3 / 0 | 9 / 0 | 8 / 0 | 2 / 0 | 3 / 0 | 369 / 3 |
| PERCENT | 33 / 0 | 241 / 3 | 203 / 0 | 14 / 0 | 51 / 0 | 42 / 0 | 62 / 0 | 129 / 1 | 23 / 0 | 12 / 0 | 810 / 4 |
| QUANTITY | 20 / 0 | 112 / 3 | 19 / 1 | 30 / 0 | 5 / 0 | 49 / 4 | 43 / 0 | 526 / 3 | 8 / 0 | 15 / 2 | 827 / 13 |
| UNIT | 2 / 20 | 62 / 107 | 4 / 18 | 6 / 29 | 3 / 5 | 4 / 48 | 5 / 44 | 35 / 478 | 4 / 8 | 0 / 16 | 125 / 773 |
| MONEY | 27 / 5 | 496 / 19 | 460 / 30 | 19 / 2 | 88 / 4 | 54 / 0 | 39 / 1 | 234 / 3 | 45 / 0 | 33 / 3 | 1495 / 67 |
| CURR | 27 / 26 | 396 / 352 | 360 / 407 | 16 / 10 | 21 / 79 | 16 / 40 | 9 / 38 | 87 / 129 | 37 / 21 | 5 / 26 | 974 / 1128 |
| Total | 8664 / 784 | 8225 / 900 | 7008 / 909 | 12097 / 939 | 6590 / 665 | 5826 / 563 | 5494 / 480 | 5961 / 788 | 8275 / 340 | 8928 / 1019 | 77068 / 7387 |

Table 9: Total number of entities in each domain in flat/nested.

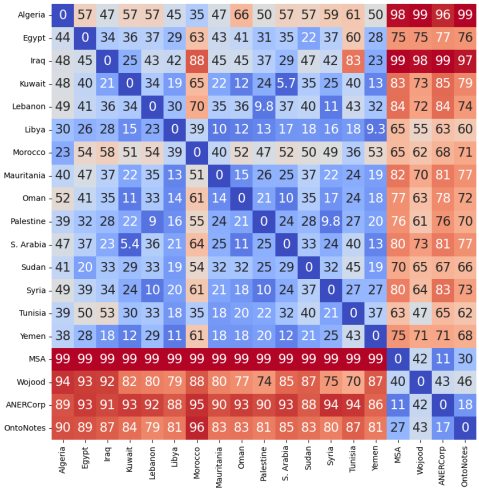

Figure 6: Heatmap showing the distances between different dialects in *Konooz* using MMD Polynomial kernel.

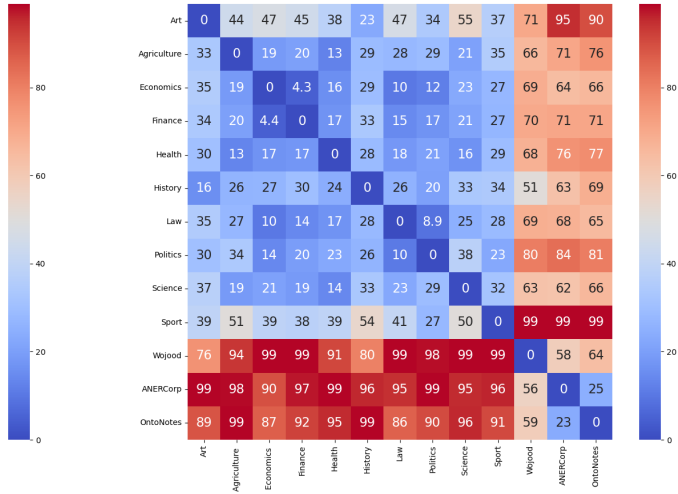

Figure 7: Heatmap showing the distances between different domains in *Konooz* using MMD Polynomial kernel.

## A.6 Training Models for Benchmarking

We trained four models, one for each dataset. Following (Obeid et al., 2020), we addressed the absence of a development set in *ANERCorp* by creating three different data splits, each including train, test, and development sets. We followed a similar approach for the *OntoNotes* model by generating three distinct data splits. For the *Wojood* dataset, we used the official splits and source code (Jarrar et al., 2022). We then fine-tuned the pre-trained models on all splits, ensuring consistency by using the same hyperparameters specified in (Obeid et al., 2020).

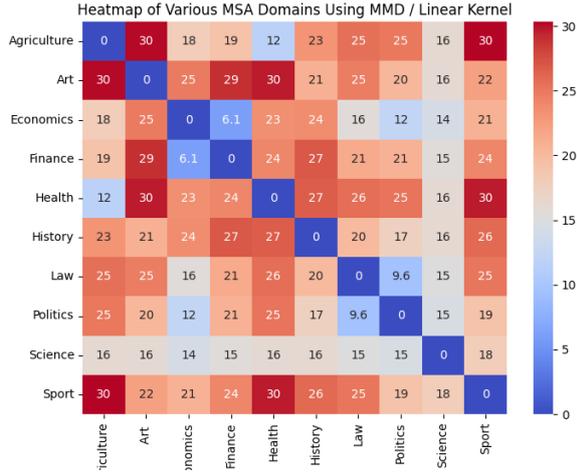

Figure 8: Heatmap showing the similarities between different MSA domains in *Konooz* using MMD Linear kernel.

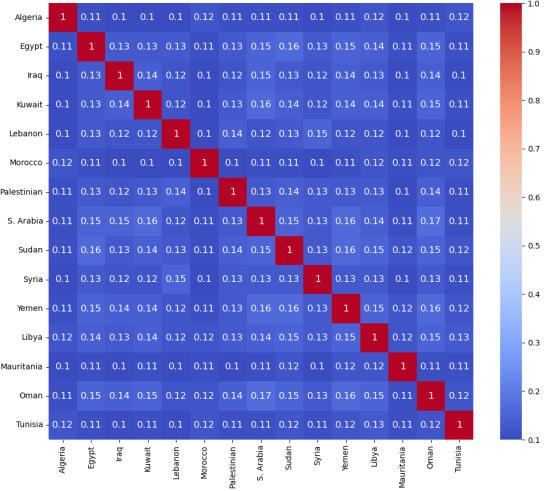

Figure 10: Heatmap showing the similarities between different dialects in *Konooz* using Jaccard coefficient.

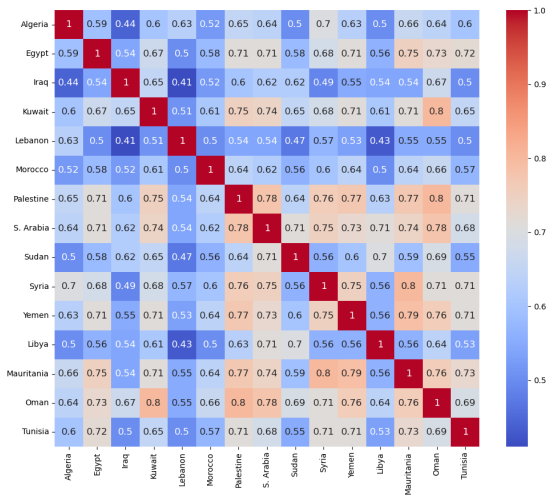

Figure 9: Heatmap showing the similarities between different dialects in *Konooz* using cosine similarity.

We also trained all datasets ($Wojood_{Nested}$, $Wojood_{Flat}$, *ANERCorp*, and *OntoNotes*) using AraBERTv1, AraBERTv2, and ArBERT. Table 10 presents the results of all datasets trained on three different pre-trained models.

All models were implemented using Hugging-Face Transformers (Wolf et al., 2020) and PyTorch (Paszke et al., 2019). We used Adam optimizer (Kingma and Ba, 2014) and a dropout rate of 0.1 (Srivastava et al., 2014). We fine-tuned the aforementioned pre-trained models using the hyperparameters presented in Table 11. Fine-tuning each model required approximately 6 hours, utilizing a system with a 1.2TB disk, 62Gi of memory, and 2 NVIDIA T4 GPUs.

After training, for each training data, we selected the best-performing split and pre-trained model based on results from the test sets.

| Model | $Wojood_{Nested}$ | $Wojood_{Flat}$ | ANERCorp | OntoNotes |
|---|---|---|---|---|
| ArBERT | **91.73** | **89.71** | 83.46 | 66.88 |
| AraBERTv2 | 91.06 | 87.3 | 79.02 | 67.33 |
| AraBERTv1 | 87.6 | 87.37 | **83.5** | **67.73** |

Table 10: The micro F1-score baseline for each model.

### A.6.1 Preparing Benchmark Datasets

Since each dataset uses a different set of entity types, we aligned *Konooz* entity tags with those of other datasets to ensure consistency. *Konooz*, adhering to the *Wojood* guidelines (Jarrar et al., 2022), includes 21 entity types, requiring no mapping. The *OntoNotes* dataset overlaps significantly but lacks certain tags, such as OCC, WEBSITE, UNIT, and CURR. In contrast, *ANERCorp* uses only four tags (PERS, ORG, LOC, and MISC). To address this, we mapped *Konooz* entity types to the corresponding tags in *OntoNotes* and *ANERCorp*, ensuring compatibility and facilitating cross-dataset analysis.

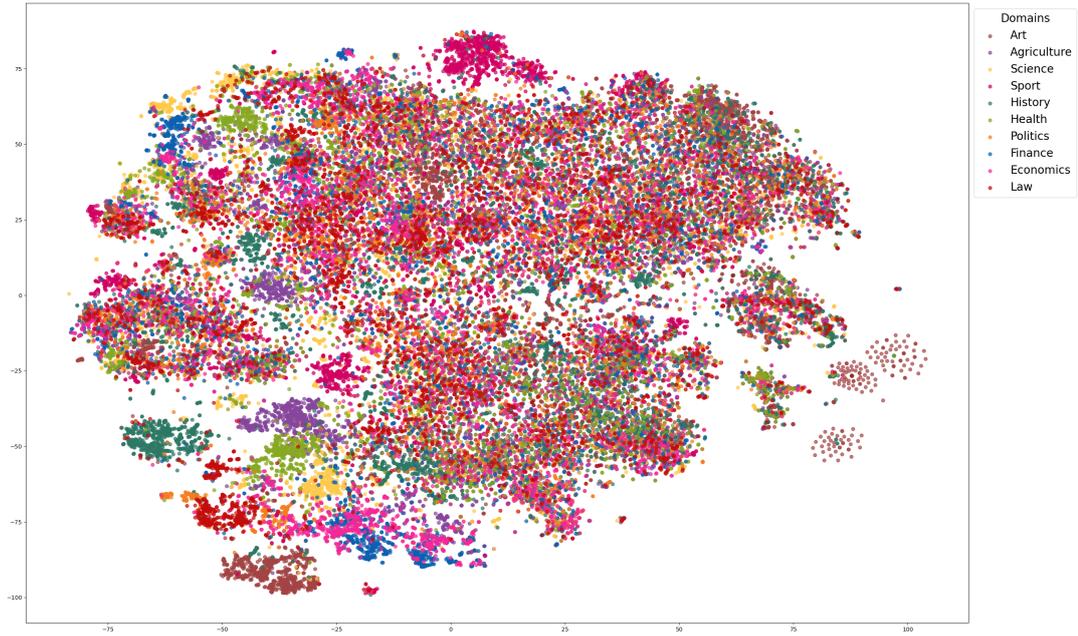

Figure 11: Lexical similarity and distribution of *Konooz* domains.

For the *ANERCorp* model, *Konooz* tags were mapped to O, except GPE and FAC, which were mapped to LOC. The MISC tag in *ANERCorp* was also mapped to O, as *Konooz* does not support it. The *OntoNotes* dataset does not support the tags OCC, WEBSITE, UNIT, and CURR, which were mapped to O in *Konooz*.

|  | Model | Batch size | LR | Epochs |
|---|---|---|---|---|
| *ANERCorp* | AraBERTv1 | 32 | $5e^{-5}$ | 3 |
| *OntoNotes* | AraBERTv1 | 8 | $5e^{-5}$ | 50 |

Table 11: Hyperparamters of best models of the *ANERCorp* and *OntoNotes* datasets.